\documentclass[10pt,twocolumn,letterpaper]{article}

\usepackage{cvpr}

\usepackage{graphicx}
\usepackage{amsmath}
\usepackage{amssymb}
\usepackage{booktabs}
\usepackage{multirow}
\usepackage{float}

\usepackage[pagebackref,breaklinks,colorlinks]{hyperref}

\usepackage[capitalize]{cleveref}
\crefname{section}{Sec.}{Secs.}
\Crefname{section}{Section}{Sections}
\Crefname{table}{Table}{Tables}
\crefname{table}{Tab.}{Tabs.}

\newcommand{\sectionskip}{1ex}
\newcommand{\subsectionskip}{1ex}
\begin{document}

\title{Unsupervised Transformer Pre-Training for Images: Self-Distillation, Mean Teachers,
    and Random Crops}

\author{Mattia Scardecchia\\
Technical University Munich\\
Arcisstraße 21, 80333 Munich, Germany\\
{\tt\small go52qab@mytum.de}
}
\maketitle


\begin{abstract}
  Recent advances in self-supervised learning (SSL) have made it possible to learn general-purpose visual features that capture both the high level semantics and the fine-grained spatial structure of images. Most notably, the recent DINOv2 has established a new state of the art by surpassing weakly supervised methods (WSL) like OpenCLIP on most benchmarks. In this survey, we examine the core ideas behind its approach, multi-crop view augmentation and self-distillation with a mean teacher, and trace their development in previous work. We then compare the performance of DINO and DINOv2 with other SSL and WSL methods across various downstream tasks, and highlight some remarkable emergent properties of their learned features with transformer backbones. We conclude by briefly discussing DINOv2's limitations, its impact, and future research directions.
\end{abstract}

\section{Introduction}
\label{sec:intro}

In the last few years, self-supervised learning (SSL) \cite{raniSelfsupervisedLearningSuccinct2023} has emerged as a powerful framework for learning representations from vast amounts of unlabeled data across various modalities, including language \cite{devlinBERTPretrainingDeep2019, brownLanguageModelsAre2020}, visual data \cite{ozbulakKnowYourSelfsupervised2023, schiappaSelfSupervisedLearningVideos2023}, audio \cite{liuAudioSelfsupervisedLearning2022}, time series \cite{zhangSelfSupervisedLearningTime2024}, and healthcare \cite{chowdhuryApplyingSelfSupervisedLearning2021}. In combination with flexible and scalable transformer-based architectures \cite{vaswaniAttentionAllYou2017}, internet-scale datasets \cite{penedoFineWebDatasetsDecanting2024, schuhmannLAION5BOpenLargescale2022}, and huge amounts of compute \cite{openaiGPT4TechnicalReport2024}, this has led to a paradigm shift in machine learning, characterized by the rise of foundation models \cite{bommasaniOpportunitiesRisksFoundation2022}. These models are trained on broad data at scale to extract robust, general-purpose representations that can easily be adapted for a wide variety of downstream tasks, without need for extensive fine-tuning and often outperforming task-specific models \cite{brownLanguageModelsAre2020, oquabDINOv2LearningRobust2024}. As a particularly striking example, large language models have shown remarkable emergent capabilities in natural language processing, question answering, and creative tasks, as well as unprecedented performance in coding and mathematics benchmarks \cite{teamGemma2Improving2024, abdinPhi3TechnicalReport2024, grattafioriLlama3Herd2024, openaiGPT4TechnicalReport2024, brownLanguageModelsAre2020, devlinBERTPretrainingDeep2019}. \\
In the domain of computer vision, for some time the most promising efforts towards replicating these successes have employed a form of textual supervision to guide visual representation learning \cite{joulinLearningVisualFeatures2015, mahajanExploringLimitsWeakly2018, radfordLearningTransferableVisual2021, chertiReproducibleScalingLaws2023}. However, these approaches require large-scale datasets of corresponding image-text pairs, which are expensive to collect. Furthermore, over-reliance on textual supervision can introduce harmful biases in the learned representations and limit their information retain, since text can only approximate the rich visual and spatial structure of images. Recently, the authors of \cite{oquabDINOv2LearningRobust2024} have shown that, leveraging discriminative SSL algorithms and curating a sufficiently large and diverse pretraining dataset, it is possible to learn general-purpose visual features from images alone, without labels nor captions. Furthermore, their method, DINOv2, learns representations that exhibit remarkable generalization capabilities across data distributions and tasks without finetuning, including on dense prediction tasks where CLIP features tend to struggle \cite{zhongRegionCLIPRegionBasedLanguageImage2022, ranasinghePerceptualGroupingContrastive2023}, surpassing weakly supervised and self-supervised alternatives alike \cite{oquabDINOv2LearningRobust2024}. \\
Motivated by these successes, in this survey we trace back the roots of DINOv2 in the literature on SSL for vision, focusing on the trends and ideas that first led to the development of its predecessor, DINO \cite{caronEmergingPropertiesSelfSupervised2021}. Indeed, DINO already displays most of the key ideas that enabled DINOv2, including framing SSL as self-distillation, using a momentum encoder as teacher, and emphasizing semantics through a multi-crop view augmentation strategy. The survey is structured as follows. First, in \cref{sec:ssl}, we provide a brief overview of key trends in SSL from images. Then, in \cref{sec:dino}, we outline the DINO algorithm and its implementation in \cite{caronEmergingPropertiesSelfSupervised2021}. In \cref{sec:backg}, we discuss the core ideas behind DINO, and trace back their development to previous work. In \cref{sec:results}, we show a quantitative comparison of DINO with previous and concurrent SSL methods on standard benchmarks, and discuss some qualitative results. Finally, in \cref{sec:extensions}, we discuss the extensions of DINO in iBOT \cite{zhouIBOTImageBERT2022} and DINOv2 \cite{oquabDINOv2LearningRobust2024}, comparing them with self-supervised and weakly supervised alternatives. We conclude by briefly discussing DINOv2's limitations, its impact, and future research directions in \cref{sec:conclusions}.

\vspace{\subsectionskip}
\subsection{Self-Supervised Learning from Images}
\label{sec:ssl}

In contrast with supervised learning, which relies on costly data annotations, self-supervised learning aims to learn useful representations from raw data, by learning to solve a pretext task in which the supervisory signal is extracted from the data itself. In computer vision, there are two important classes of such methods. The first type extracts the error signal from individual images, by transforming, corrupting, or masking parts of the input, and learning to recognize the transformation or recover the information that was lost. Early methods of this type often exploited prior knowledge about the visual and spatial structure of images to define meaningful surrogate tasks, like image colorization \cite{zhangColorfulImageColorization2016}, rotation prediction \cite{gidarisUnsupervisedRepresentationLearning2018}, jigsaw puzzles \cite{norooziUnsupervisedLearningVisual2017}, or inpainting \cite{pathakContextEncodersFeature2016}. Generative methods for SSL \cite{vincentStackedDenoisingAutoencodersa, higginsBetaVAELearningBasic2016, heMaskedAutoencodersAre2021, baoBEiTBERTPreTraining2022} can be regarded as a subclass that makes minimal assumptions about the structure of data, learning to reconstruct the input from a corrupted version of it. While this makes them more flexible and easily generalizable to different modalities \cite{huangMaskedAutoencodersThat2022, tongVideoMAEMaskedAutoencoders2022, girdharOmniMAESingleModel2023}, they tend to learn representations of lower semantic level compared to discriminative methods, and usually require full finetuning to achieve competitive performance with discriminative methods on downstream tasks \cite{heMaskedAutoencodersAre2021, assranSelfSupervisedLearningImages2023}. \\ 
A second class of methods create the supervisory signal by encouraging an encoder to learn embeddings that discriminate between images or groups of images, and are therefore often referred to as discriminative methods \cite{dosovitskiyDiscriminativeUnsupervisedFeature2014, wuUnsupervisedFeatureLearning2018, caronDeepClusteringUnsupervised2019, chenSimpleFrameworkContrastive2020, grillBootstrapYourOwn2020, zbontarBarlowTwinsSelfSupervised2021}. Many successful discriminative algorithms leverage a \textit{joint embedding} framework, where an encoder is trained to output similar representations for different \textit{views} of the same input, generated through hand-crafted data augmentations. These methods are susceptible to a phenomenon called representation collapse, where the encoder learns a trivial mapping that solves the task but fails to capture anything meaningful about the underlying data structure \cite{balestrieroCookbookSelfSupervisedLearning2023}. To prevent this from happening, they leverage various expedients that can be thought of as maximizing the volume of feature space occupied by the embeddings \cite{assranHiddenUniformCluster2022}, such as using contrasting negative examples \cite{chenSimpleFrameworkContrastive2020, heMomentumContrastUnsupervised2020}, removing correlations in feature space \cite{bardesVICREGVARIANCEINVARIANCECOVARIANCERE2022, zbontarBarlowTwinsSelfSupervised2021}, or discriminating between high-entropy clusterings of the data \cite{caronDeepClusteringUnsupervised2019, asanoSelflabellingSimultaneousClustering2020}. Other discriminative methods use a \textit{joint embedding predictive} architecture, with two encoders, being fed different views of the same inputs, that now play an asymmetric role: a predictor network must learn to map the embeddings of the \textit{online encoder} into those of the \textit{target encoder} \cite{grillBootstrapYourOwn2020, chenExploringSimpleSiamese2020}. In these methods, the asymmetry due to the predictor is key to avoid representation collapse, as long as gradients are prevented from flowing directly through the target encoder \cite{chenExploringSimpleSiamese2020}. An informal argument for why this is the case, assuming an optimal predictor, can be found in \cite{grillBootstrapYourOwn2020}. \\ 
DINO \cite{caronEmergingPropertiesSelfSupervised2021}, the focus of this survey, is most closely related with methods in this last group, especially BYOL \cite{grillBootstrapYourOwn2020} and SimSiam \cite{chenExploringSimpleSiamese2020}, which learn by bootstrapping their own objective. Differently from them, however, it removes the predictor and avoids collapse through a simple centering and sharpening of the target embeddings. This simplification is what allows to interpret DINO as a form of self-\textbf{di}stillation with \textbf{no} labels \cite{caronEmergingPropertiesSelfSupervised2021}. We discuss it further in \cref{sec:backg}.

\vspace{\sectionskip}
\section{DINO}
\label{sec:dino}

\begin{figure*}[t]
    \centering
    \includegraphics[width=\linewidth]{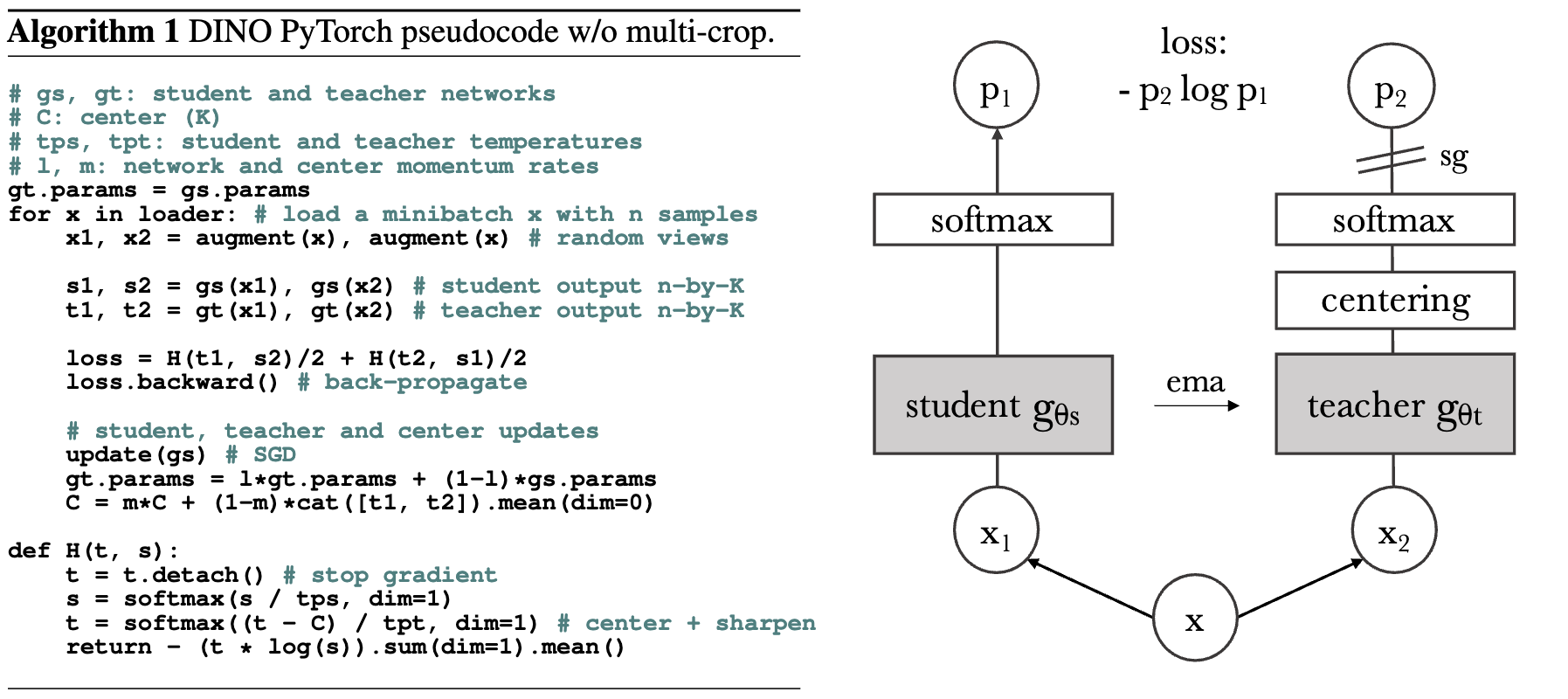}
    \caption{\textbf{DINO algorithm without multi-crop.} Two views of the same input are processed by student and teacher encoders, sharing same architecture but different parameters. The teacher output is centered using batch statistics, then both outputs are normalized with a temperature softmax. Teacher weights are an EMA of the student's. Embeddings similarity is computed as a cross-entropy. (Left) PyTorch pseudocode of DINO. (Right) Diagram of DINO. Figures from \cite{caronEmergingPropertiesSelfSupervised2021}.}
    \label{fig:dino-algo}
\end{figure*}

To set the stage for the discussion in the next section, here we describe the DINO algorithm and its implementation in \cite{caronEmergingPropertiesSelfSupervised2021}.

\vspace{\subsectionskip}
\subsection{Algorithm}

DINO is a joint embedding method, with a student encoder \(g_{\theta_s}\) and a teacher encoder \(g_{\theta_t}\) which share the same architecture (\cref{fig:dino-algo}). The teacher's weights are maintained as an exponential moving average (EMA) of the student's weights, like in BYOL \cite{grillBootstrapYourOwn2020}, while the student is trained to match the teacher's outputs when provided with different augmented views of the same image. \\
Views are generated through the multi-crop strategy introduced in \cite{caronUnsupervisedLearningVisual2021}: from a single image, we sample two \textit{global} crops and several \textit{local} crops of smaller resolution. Each crop is augmented independently following BYOL \cite{grillBootstrapYourOwn2020} (color jittering, Gaussian blur, solarization), giving a set \(V\) containing two global views \(x_1^g\), \(x_2^g\) and several local views. The student processes all views, while the teacher processes only the global ones, thus encouraging semantically rich 'local-to-global' correspondences. \\
The teacher's outputs are centered using a mean embedding, maintained as an EMA across subsequent batches, as can be seen in the pseudocode in \cref{fig:dino-algo}. Then, both teacher and student outputs are normalized through a temperature softmax, using a low temperature for the teacher to achieve target 'sharpening'. If \(P_s(\cdot)\) and \(P_t(\cdot)\) denote the full transformations from image view to embedding just described, and \(H(\cdot, \cdot)\) is the cross-entropy loss, the student loss is computed as: 
\begin{equation}
    \mathcal{L}(\theta_s) = \sum_{x \in \{x^{g}_1, x^{g}_2\}} \quad \sum_{\substack{x' \in V \setminus \{x\}}} \quad H(P_t(x), P_s(x'))
\label{eq:loss}
\end{equation}

\subsection{Encoder Architecture}

The architecture of the DINO encoders is composite: it consists of a backbone, which in \cite{caronEmergingPropertiesSelfSupervised2021} is either a ResNet \cite{heDeepResidualLearning2015} or a Vision Transformer (ViT) \cite{dosovitskiyIMAGEWORTH16X162021}, followed by a projection head. This design was introduced by SimCLR \cite{chenSimpleFrameworkContrastive2020} for contrastive learning, and was later adopted by SwAV \cite{caronUnsupervisedLearningVisual2021} and BYOL \cite{grillBootstrapYourOwn2020} as well. The reason for this choice has to do with a phenomenon known as dimensional collapse \cite{huaFeatureDecorrelationSelfSupervised2021, jingUnderstandingDimensionalCollapse2022}, which is common in many joint embedding methods \cite{balestrieroCookbookSelfSupervisedLearning2023}. \\ 
Dimensional collapse occurs when the information encoded in different dimensions of the embeddings is redundant, which can be diagnosed by inspecting their spectrum \cite{balestrieroCookbookSelfSupervisedLearning2023}. The basic mechanism is the following: the task encourages the encoder to learn representations that are invariant to view augmentations; without a strong enough volume maximization constraint \cite{assranHiddenUniformCluster2022}, or other counteracting measures, the encoder will tend to collapse its representations to have an easier time solving the task \cite{chenSimpleFrameworkContrastive2020}. \\
Surprisingly, even in contrastive methods like SimCLR, dimensional collapse is observed \cite{chenSimpleFrameworkContrastive2020}. The authors of \cite{jingUnderstandingDimensionalCollapse2022} have proposed two possible explanations for this. On one hand, if data augmentation is very strong, it might induce a variance in some data feature that is comparable or stronger than the natural variance of the dataset along that direction, swamping the discriminative signal. This would incentivize the encoder to become insensitive to that feature, leading to collapse \cite{jingUnderstandingDimensionalCollapse2022}. On the other hand, with weaker augmentation, the implicit regularization of over-parametrized neural networks might be responsible. Indeed, the authors of \cite{jingUnderstandingDimensionalCollapse2022} argue that these networks tend to find low rank solutions, which can prevent them from encoding more than minimal information, leading to collapse. \\
With all this in mind, the rationale for the composite design becomes clear: we hope to extract representations from the network just before the collapse happens. And in fact, empirically, introducing a learnable non-linear transformation between the backbone and the loss computation has been found to improve downstream performance and mitigate collapse \cite{chenSimpleFrameworkContrastive2020, caronUnsupervisedLearningVisual2021, grillBootstrapYourOwn2020}. In DINO, the projection head is composition of a 3-layer MLP, \(l_2\) normalization, and a weight normalized fully-connected layer \cite{salimansWeightNormalizationSimple2016}. As for the backbone, when a ViT is used, the image embedding is taken to be the embedding of the [CLS] token. This is not associated with any literal 'class', but it interacts with the patch embeddings through the self-attention layers \cite{vaswaniAttentionAllYou2017}, and learns a global embedding for the image guided by the self-distillation objective, playing a similar role to the [CLS] token in BERT \cite{devlinBERTPretrainingDeep2019}.

\vspace{\sectionskip}
\section{Ideas behind DINO}
\label{sec:backg}

\subsection{View Augmentation and multi-crop}

\begin{figure*}[t]
  \centering
  \includegraphics[width=\linewidth]{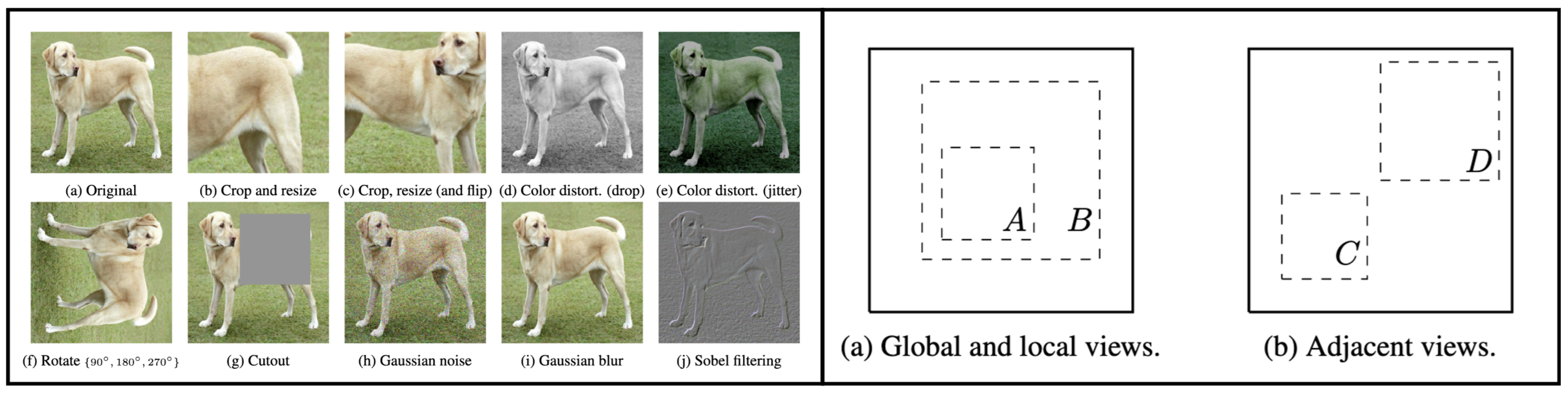}
  \caption{\textbf{View Augmentation in discriminative SSL for images}. (Left): Illustration of common stochastic data augmentation operations used for view augmentation. (Right): Random cropping generates semantically rich view correspondences, including adjacency and global-local relationships. Figures from \cite{chenSimpleFrameworkContrastive2020}.}
  \label{fig:view-augmentation}
\end{figure*}


View augmentation is a key component of most discriminative SSL methods for vision, including DINO \cite{dosovitskiyDiscriminativeUnsupervisedFeature2014, wuUnsupervisedFeatureLearning2018, bachmanLearningRepresentationsMaximizing2019, chenSimpleFrameworkContrastive2020, heMomentumContrastUnsupervised2020, chenExploringSimpleSiamese2020, grillBootstrapYourOwn2020, caronUnsupervisedLearningVisual2021, caronEmergingPropertiesSelfSupervised2021}. It is a form of stochastic data augmentation that generates multiple views of the same input, and it is used to induce invariance to a class of 'style' transformations in the learned representations, while making them discriminative of the semantic content of images. The choice of augmentation strategy relies on prior knowledge about the input modality, and can significantly affect the quality of the learned representations \cite{chenSimpleFrameworkContrastive2020}. \\
Some early discriminative methods that crucially relied on this principle framed the SSL problem as a classification task, treating each sample in a dataset as its own class, up to view augmentation \cite{dosovitskiyDiscriminativeUnsupervisedFeature2014, wuUnsupervisedFeatureLearning2018}. For example, in \cite{dosovitskiyDiscriminativeUnsupervisedFeature2014} the authors used a concatenation of random elementary transformations (random cropping, rotation, translation, scaling, contrast and color distortions) to generate a number of augmented views from each image in a dataset, and then trained a convolutional neural network to recognize augmented crops coming from the same source image. They framed this explicitly as a classification task: if \(x\) is an image view and \(v = f_\theta(x)\) its features just before the softmax, the probability of \(x\) being recognized as the i-th sample is:
\begin{equation}
    p(i|v) = \frac{\exp(w_i^Tv)}{\sum_j \exp(w_j^Tv)}
\end{equation}
Later, the authors of \cite{wuUnsupervisedFeatureLearning2018} proposed a non-parametric formulation of the same classification task, where the class probability is computed as:
\begin{equation}
  p(i|v) = \frac{\exp(v_i^Tv/\tau)}{\sum_j \exp(v_j^Tv/\tau)}
\end{equation}
Here, the learnable class prototypes \(w_j\) are replaced with the l2-normalized features \(v_j\) of all samples, extracted by the network at previous iterations during training, and retrieved from a memory bank. This metric learning formulation, pushing the normalized embeddings to maximally scatter on a hypersphere, relies on direct comparisons between embeddings and  has been shown to greatly improve the quality of the learned representations \cite{wuUnsupervisedFeatureLearning2018}. To scale their approach to large datasets, the authors used Noise-Contrastive Estimation \cite{maNoiseContrastiveEstimation2018} to efficiently approximate the non-parametric softmax computation. Furthermore, they introduced a proximal regularization term encouraging class features to be consistent across epochs, which plays a similar stabilizing role to the momentum encoder in subsequent methods like MoCo \cite{heMomentumContrastUnsupervised2020}. \\
Many recent methods in the instance discrimination family have moved away from an explicit classification formulation, and rather resort to contrastive learning to directly learn an encoder whose embeddings can discriminate different samples, while being invariant to some transformations. Most of them rely on a contrastive loss \cite{bromleySignatureVerificationUsing1993, chopraLearningSimilarityMetric2005, hadsellDimensionalityReductionLearning2006} to encourage the embeddings of 'positive' pairs of inputs to be close, while pushing the embeddings of 'negative' pairs apart. A popular choice in SSL in the infoNCE loss \cite{ohsongDeepMetricLearning2016, oordRepresentationLearningContrastive2019}:
\begin{equation}
    \mathcal{L} = -\sum_{(i, j)} \log\frac{\exp(sim(r_i, r_j) / \tau)}{\sum_{k=1}^{N}\exp(sim(r_i, r_k) / \tau)}
\end{equation}
where we are given \(N\) inputs with representations \(r_1, ..., r_N\), and the external sum is over positive pairs \( (i, j) \). The similarity function \(sim\) is typically the cosine similarity, and the temperature \(\tau\) controls the concentration of the softmax. A particularly influential work in this direction is SimCLR \cite{chenSimpleFrameworkContrastive2020}, which first introduced a simple approach to contrastive learning based on view augmentation without need for specialized architectures \cite{bachmanLearningRepresentationsMaximizing2019, henaffDataEfficientImageRecognition2020}, nor a memory bank \cite{wuUnsupervisedFeatureLearning2018, tianContrastiveMultiviewCoding2020, heMomentumContrastUnsupervised2020, misraSelfSupervisedLearningPretextInvariant2020}. They use a joint embedding predictive architecture with two copies of the same encoder, and a contrastive loss that encourages the embeddings of different augmented views of the same image to be close. As negative examples, they employ the augmented views of other images in the batch. With this simple design, view augmentation becomes a flexible way to define the contrastive predictive task \cite{chenSimpleFrameworkContrastive2020}. Key to the success of their method is the use of random crops in the augmentation pipeline, applied before random flips, color distortions, and Gaussian blur. Indeed, this generates a semantically rich set of view pairs, exhibiting both adjacency and local-global correspondences (\cref{fig:view-augmentation}), leading to stronger visual representations \cite{chenSimpleFrameworkContrastive2020}. \\
The view augmentation strategy of SimCLR was further refined in SwAV \cite{caronUnsupervisedLearningVisual2021}, a clustering-based discriminative method \cite{caronDeepClusteringUnsupervised2019, asanoSelflabellingSimultaneousClustering2020, caronUnsupervisedLearningVisual2021} that uses a joint embedding architecture with two copies of the same encoder. Instead of comparing directly the embeddings of two views, in SwAV the features are projected onto a set of learnable prototype vectors (centroids, sort of), obtaining 'codes'. Then, the codes obtained from each augmented view are predicted from the embedding of the other. A key innovation, decoupled from the architecture, is the introduction of a multi-crop strategy for view augmentation: for each image, they generate two global views at standard resolution, and several local views at lower resolution, all of which are then augmented independently. Codes are computed only with the two global views, and each code is predicted from the embeddings of all other views: if \(z_1, z_2\) are the embeddings of the two global views, \(q_1, q_2\) their codes, and \(z_3, \dots, z_{V+2}\) are the embeddings of the local views, the loss is computed as
\begin{equation}
  L(\mathbf{z}_{1}, \mathbf{z}_{2}, \dots, \mathbf{z}_{V + 2}) = \sum_{i \in \{1, 2\}} \sum_{v=1}^{V + 2} \mathbf{1}_{v \neq i} \cdot \ell(\mathbf{z}_{v}, \mathbf{q}_{i}).
\end{equation}
For a small increase in computational cost, this allows to compare a much larger number of view pairs per image, and it encourages the model to learn semantically rich global-local correspondences. As a consequence, multi-crop significantly improves the downstream performance of several discriminative methods, including SwAV, SimCLR, DeepCluster, and SeLa \cite{caronUnsupervisedLearningVisual2021, chenSimpleFrameworkContrastive2020, caronDeepClusteringUnsupervised2019, asanoSelflabellingSimultaneousClustering2020}, and it will be important in DINO as well.

\vspace{\subsectionskip}
\subsection{Self-Distillation}

DINO can be seen as a form of self-distillation, where the teacher is built from previous iterations of the student network and shares its same architecture. Originally, knowledge distillation was proposed as a means to compress the knowledge of an already trained ensemble of models into a single one, to achieve compression and efficiency gains \cite{buciluaModelCompression2006, hintonDistillingKnowledgeNeural2015}. For example, in the context of classification, the authors of  \cite{hintonDistillingKnowledgeNeural2015} trained a student to minimize the cross-entropy between its predicted probabilities and those output by an ensemble as teacher. They included a temperature in the teacher softmax, which plays an important role: while the non-maximum logits output by the teacher contain the 'dark knowledge' acquired by the ensemble, the most negative ones are under-constrained and hence less reliable. Tuning the temperature allows to strike the right balance along this axis, significantly outperforming a student trained from scratch without distillation \cite{hintonDistillingKnowledgeNeural2015}. \\
Perhaps more surprisingly, the authors of \cite{furlanelloBornAgainNeural2018} showed that a student sharing the same architecture as a single teacher model can significantly outperform it if trained to imitate it using knowledge distillation. This procedure is often called 'self-distillation', but to avoid ambiguities with self-distillation in SSL we will refer to it with the non-standard name 'twin-knowledge-distillation'. Later, the authors of \cite{allen-zhuUnderstandingEnsembleKnowledge2023} derived theoretical results that shed light on the efficacy of ensembling, knowledge distillation, and twin-knowledge-distillation in deep learning, focusing on classification. Their key insight is that these techniques are provably effective when the data exhibits a 'multi-view' structure: in a sense that can be made precise, multi-view data exhibits multiple independent features which can all be used to correctly infer the label. Assuming this redundant structure in the data (which is common in practice, e.g. in natural images), they show that an ensemble of independently trained neural networks can improve test accuracy, and that this superior performance can be distilled into a single model with a procedure similar to that of \cite{hintonDistillingKnowledgeNeural2015}. Furthermore, they prove that twin-knowledge-distillation can be viewed as implicitly combining ensembling of two models and distillation of the knowledge of the ensemble into a single one, thus explaining the boost in performance \cite{allen-zhuUnderstandingEnsembleKnowledge2023}. \\
Building on these ideas, the authors of \cite{xieSelftrainingNoisyStudent2020} combined twin-knowledge-distillation with self-training, an approach for semi-supervised learning that aims at propagating a small initial set of annotations to a much larger corpus of unlabeled samples, by using a partially trained model to generate pseudo-labels \cite{aminiSelfTrainingSurvey2024}. By iteratively training an equal-or-larger student with the labels generated by the previous student model as teacher, and injecting strong noise in the student's training through dropout, stochastic depth, and data augmentation, they show that each student can leverage unlabelled data to learn to generalize better than its teacher for many generations \cite{xieSelftrainingNoisyStudent2020}. Another step forward towards self-distillation as in DINO is the work of \cite{zhangBeYourOwn2019}, which was the first to introduce the idea of performing a form of knowledge distillation 'online', with the model acting as its own teacher during training. They divide a neural network classifier into subsequent chunks and append a bottlenecked classification head at the end of each, training each head to perform the task. At the same time, they distill knowledge from deeper layers to shallower ones: each chunk is trained to mimic the output of the deeper one, minimizing a KL divergence between predicted distributions and a \(l2\) distance between internal representations in the bottlenecks. Interestingly, their method finds solutions lying in flatter regions of the loss landscape, and it significantly outperforms a baseline obtained without their proposed 'self-distillation' \cite{zhangBeYourOwn2019}. \\
In the context of SSL, BYOL \cite{grillBootstrapYourOwn2020} is very close to the self-distillation approach of DINO. It uses a joint embedding predictive architecture, where online and target encoders share the same composite architecture \cite{chenSimpleFrameworkContrastive2020}, and the target encoder is updated as an EMA of the online encoder. Unlike contrastive methods, BYOL does not use negative samples: it bootstraps an objective using the target encoder as a teacher, and trains the online encoder + predictor to match its outputs, as measured by the mean squared error. DINO's teacher-student formulation is extremely similar to BYOL's, with the most significant difference being the way in which collapse is avoided (predictor in BYOL, centering and sharpening of targets in DINO).

\vspace{\subsectionskip}
\subsection{Mean Teachers}

\begin{figure*}[t]
  \centering
  \includegraphics[width=\linewidth]{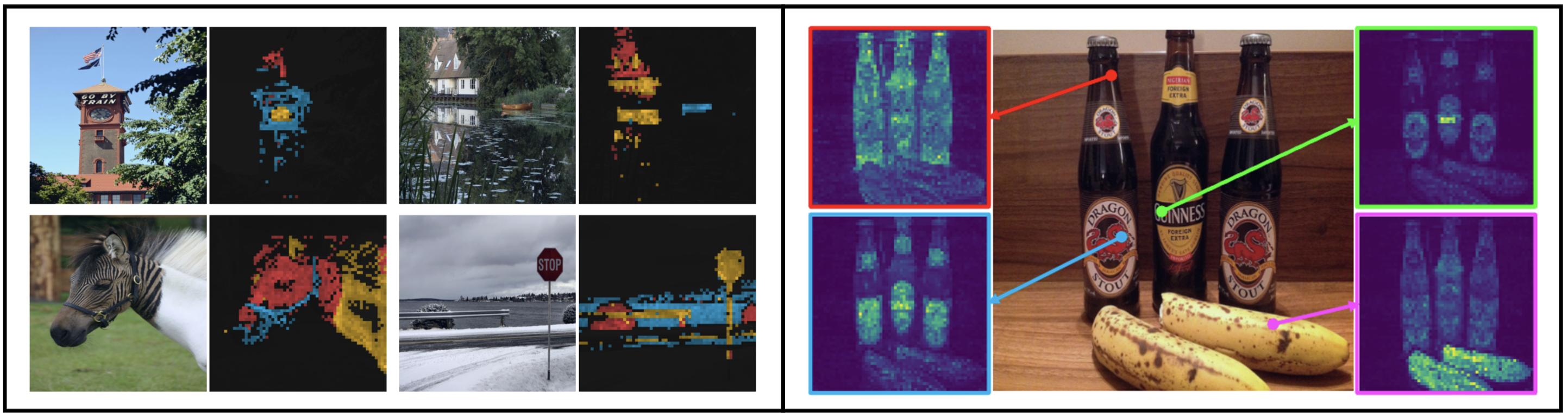}
  \caption{\textbf{Self-attention patterns of a ViT trained with DINO}. Visualization of the self-attention weights in the last layer of a ViT-S trained with DINO. (Left): Response to the query of the [CLS] token, with different heads encoded using different colors. Each head focuses on different objects or parts (Right): Responses to the queries of several patch tokens. The network has learned to separate objects. Figures from \cite{caronEmergingPropertiesSelfSupervised2021}.}
  \label{fig:dino-attention-fine}
\end{figure*}

The teacher in DINO is maintained as an exponential moving average of the student's weights. This choice is related to several previous works, from various angles. The use of a momentum encoder in discriminative SSL was first introduced by MoCo \cite{heMomentumContrastUnsupervised2020, chenImprovedBaselinesMomentum2020, chenEmpiricalStudyTraining2021}, which proposed a general scheme for contrastive pretraining with view augmentations. Contrastive methods require a large numer of negative samples to mine sufficiently hard ones by chance. In methods like SimCLR, which use other elements in the batch as negative samples, the batch size is coupled with the 'dictionary size' of negative samples, forcing the use of large batch sizes to achieve good performance \cite{chenSimpleFrameworkContrastive2020}. To decouple the two, a memory bank as in \cite{wuUnsupervisedFeatureLearning2018} could be used, but this would introduce a discrepancy between actual current embeddings and those stored in the memory bank, especially with large datasets. MoCo solves this problem by using a momentum encoder to embed negative samples from a queue on the fly, maintaining embedding consistency while allowing training with small batch sizes \cite{heMomentumContrastUnsupervised2020}. \\
Closer to DINO, the already introduced BYOL \cite{grillBootstrapYourOwn2020} uses a momentum encoder as a teacher to provide more stable targets for its bootstrapped objective. While initially this was thought to be necessary to avoid representation collapse, a more recent version of the BYOL paper showed that this is not the case, and this finding was confirmed by SimSiam \cite{chenExploringSimpleSiamese2020}: only the stop gradient operation is necessary to avoid collapse, but the use of a momentum encoder can produce more stable targets and higher quality representations \cite{chenExploringSimpleSiamese2020}. In this sense, the use of EMA updates for the teacher in BYOL and DINO is similar to the use of EMA updates for target networks \cite{mnihPlayingAtariDeep2013} in Deep Reinforcement Learning, and especially with actor-critic methods. Indeed, many such methods are trained with a bootstrapped objective derived from Bellman optimality equations, and the introduction of a target network is crucial to stabilize training and achieve good performance \cite{mnihPlayingAtariDeep2013, haarnojaSoftActorCriticOffPolicy2018, lillicrapContinuousControlDeep2019}. While initially target networks were updated by regularly copying the weights of the online network \cite{mnihPlayingAtariDeep2013}, more recent methods found it beneficial to use EMA updates instead, like for the target critic in SAC \cite{haarnojaSoftActorCriticOffPolicy2018}, or the target actor and critic in DDPG \cite{lillicrapContinuousControlDeep2019}. \\
Also related to DINO's EMA teacher is the use of a mean teacher in semi-supervised learning, as in \cite{tarvainenMeanTeachersAre2018}. A class of semi-supervised methods, inspired by \cite{rasmusSemiSupervisedLearningLadder2015}, introduce a consistency cost that encourages the model to produce similar outputs on different noisy versions of the same input, to exploit unlabelled data. In this context, \cite{laineTemporalEnsemblingSemiSupervised2017} had proposed to maintain an EMA of model predictions across epochs for each sample, to use them as targets in the consistency cost. Building on this approach, the authors of \cite{tarvainenMeanTeachersAre2018} proposed to instead use a mean teacher, whose weights are an EMA of the model's weights during training, to generate such targets on the fly. This approach accelerates the rate at which new knowledge is incorporated in the consistency objective, especially with large datasets, and allows extending to the online setting. Furthermore, it improves the final performance compared to averaging outputs \cite{tarvainenMeanTeachersAre2018}. While significant, this improvement is not entirely surprising: it is well known that averaging subsequent versions of a model along an optimization trajectory in weight space tends to produce configurations lying in flatter regions of the loss landscape, associated with better generalization performance \cite{chaudhariEntropySGDBiasingGradient2017, izmailovAveragingWeightsLeads2019}. And in fact also in DINO, throughout training, the representations of the EMA teacher consistently exhibit superior downstream performance compared to those of the student (\cref{fig:dino-diagnostics}) \cite{caronEmergingPropertiesSelfSupervised2021}.

\vspace{\sectionskip}
\section{Results}
\label{sec:results}

In this section, we discuss how DINO \cite{caronEmergingPropertiesSelfSupervised2021} compares with previous and concurrent SSL methods in terms of downstream performance, and we discuss some qualitative features of the representations learned with DINO, especially using a transformer backbone.

\vspace{\subsectionskip}
\subsection{Downstream Performance}
\label{sec:dino-downstream}

Typically, to evaluate SSL methods, we measure their performance under different transfer learning protocols on a variety of donwstream tasks. The specification of such benchmarks can vary along several orthogonal axes, including the adaptation method (e.g., full fine-tuning, linear probing, few-shot fine-tuning, zero-shot feature extraction, etc.), the similarity of the pretraining and downstream data distributions (e.g., in-domain, out-of-domain), and the nature of the downstream task \cite{marksCloserLookBenchmarking2024}. In computer vision, commonly considered tasks include image classification, object detection, semantic segmentation, depth estimation, and image retrieval \cite{marksCloserLookBenchmarking2024}. \\
To obtain a wide perspective on the comparison between DINO and other SSL methods, in \cref{tab:dino-imagenet} we summarized the transfer performance of several previous and concurrent methods, all pretrained on ImageNet-1k \cite{russakovskyImageNetLargeScale2015}, on the eval split of the same dataset. This has been a de-facto standard evaluation protocol for a long time, and it allows to include representatives of the main classes of SSL methods in our comparison, such as generative \cite{assranMaskedSiameseNetworks2022, baoBEiTBERTPreTraining2022}, contrastive \cite{heMomentumContrastUnsupervised2020, chenSimpleFrameworkContrastive2020}, clustering-based \cite{caronDeepClusteringUnsupervised2019, caronUnsupervisedLearningVisual2021}, distillation-based \cite{chenExploringSimpleSiamese2020, grillBootstrapYourOwn2020}, and info-max methods \cite{zbontarBarlowTwinsSelfSupervised2021, bardesVICREGVARIANCEINVARIANCECOVARIANCERE2022}. The comparison is stratified by backbone architecture, and considers linear probing, kNN, and full finetuning protocols, where available. For each architecture we also include the accuracy obtained through supervised learning with the method of \cite{touvronTrainingDataefficientImage2021}. Reported metrics are mostly taken from the original papers introducing each method, or in some cases from re-implementations by the authors of DINO \cite{caronEmergingPropertiesSelfSupervised2021}. Furthermore, to facilitate the reader in navigating the complex SSL landscape, we included some extra information about each method, including whether it's generative or discriminative, how it prevents collapse, whether it uses a memory bank or a momentum encoder, and whether it can tolerate small batch sizes. \\
There are three main takeaways from the comparison in \cref{tab:dino-imagenet}. First, with a ResNet-50 backbone, DINO matches the SOTA-at-the-time with linear probing, and slightly surpasses it with kNN evaluation. This is remarkable considering the relative simplicity of the method, which dispenses with the need for negative samples and employs a simple joint embedding architecture. Second, with a ViT-S backbone, DINO outperforms other methods by a large margin in off-the-shelf evaluations, with a 3.5\% improvement under linear probing and an impressive 7.9\% improvement under kNN evaluation. According to ablations in \cite{caronEmergingPropertiesSelfSupervised2021}, this crucially depends on the combination of momentum encoder and multi-crop augmentation, and only emerges with a ViT backbone. Third, generative methods like MAE and BEiT struggle with off-the-shelf evaluations, but fully catch up with DINO in full finetuning, confirming the widely held belief that they tend to learn representations of a lower semantic level \cite{assranSelfSupervisedLearningImages2023}. Crucially, we make this observation controlling for the backbone architecture, which has been found to be a confounder in this respect by the authors of \cite{marksCloserLookBenchmarking2024}. \\
The authors of \cite{caronEmergingPropertiesSelfSupervised2021} investigated the quality of frozen DINO features on several other benchmarks, including image retrieval, copy detection, and video segmentation. They found that, without any finetuning, DINO achieves better performance than supervised baselines on these tasks, and is often competitive with specialized models \cite{caronEmergingPropertiesSelfSupervised2021}. This underscores the quality of DINO's representations as off-the-shelf features. The strong performance on video segmentation (DAVIS 2017) is particularly remarkable, since it indicates that the model's representations retain fine-grained spatial information about images, despite no explicit incentive to do so in the training objective \cite{caronEmergingPropertiesSelfSupervised2021}. In \cref{tab:vision_models} (described in \cref{sec:extensions}), we report metrics for DINO on a variety of benchmarks, comparing with more recent approaches.


\begin{table*}[t]
  \centering
  \scriptsize
  \label{tab:dino-imagenet}
  \begin{tabular}{l c c c c c c c c c c c c}
    \toprule
    Method
    & Discr & Gen & Contr & Distil & Clust & Info & ME 
    & MB & small BS
    & \multicolumn{3}{c}{INet-1k} \\
    \cmidrule(lr){11-13}
    & & & & & & & & & & linear & kNN & full ft \\
    \midrule
    \multicolumn{13}{c}{\textbf{ResNet-50} (23M params, 1237 im/s)} \\
    \midrule
    Supervised \cite{touvronTrainingDataefficientImage2021}
      & -- & -- & -- & -- & -- & -- & -- 
      & \texttimes & \checkmark
      & 79.3 & 79.3 & 79.3 \\ 
    Exemplar \cite{dosovitskiyDiscriminativeUnsupervisedFeature2014, doerschMultiTaskSelfSupervisedVisual2017}
      & \checkmark & \texttimes & \texttimes & \texttimes & \texttimes & \texttimes & \texttimes
      & \texttimes & \checkmark
      & 31.5 & -- & -- \\ 
    InstDiscr \cite{wuUnsupervisedFeatureLearning2018}
      & \checkmark & \texttimes & \texttimes & \texttimes & \texttimes & \texttimes & \texttimes
      & \checkmark & \checkmark
      & 54.0 & 46.5 & -- \\ 
    MoCo \cite{heMomentumContrastUnsupervised2020}
      & \checkmark & \texttimes & \checkmark & \texttimes & \texttimes & \texttimes & \checkmark
      & \checkmark & \checkmark
      & 60.6 & -- & -- \\ 
    PIRL \cite{misraSelfSupervisedLearningPretextInvariant2020}
      & \checkmark & \texttimes & \checkmark & \texttimes & \texttimes & \texttimes & \texttimes
      & \checkmark & \checkmark
      & 63.6 & -- & -- \\ 
    CPCv2 \cite{henaffDataEfficientImageRecognition2020}
      & \checkmark & \texttimes & \checkmark & \texttimes & \texttimes & \texttimes & \texttimes
      & \texttimes & \texttimes
      & 63.8 & -- & -- \\ 
    SimCLR \cite{chenSimpleFrameworkContrastive2020}
      & \checkmark & \texttimes & \checkmark & \texttimes & \texttimes & \texttimes & \texttimes
      & \texttimes & \texttimes
      & 69.1 & 60.7 & -- \\ 
    MoCov2 \cite{chenImprovedBaselinesMomentum2020, caronEmergingPropertiesSelfSupervised2021}
      & \checkmark & \texttimes & \checkmark & \texttimes & \texttimes & \texttimes & \checkmark
      & \checkmark & \checkmark
      & 71.1 & 61.9 & -- \\ 
    SimCLRv2 \cite{chenBigSelfSupervisedModels2020a}
      & \checkmark & \texttimes & \checkmark & \texttimes & \texttimes & \texttimes & \texttimes
      & \checkmark & \checkmark
      & 71.7 & -- & -- \\ 
    BarlowT \cite{zbontarBarlowTwinsSelfSupervised2021}
      & \checkmark & \texttimes & \texttimes & \texttimes & \texttimes & \checkmark & \texttimes
      & \texttimes & \checkmark
      & 73.2 & 66.0 & -- \\ 
    VICReg \cite{bardesVICREGVARIANCEINVARIANCECOVARIANCERE2022}
      & \checkmark & \texttimes & \texttimes & \texttimes & \texttimes & \checkmark & \texttimes
      & \texttimes & \checkmark
      & 73.2 & -- & -- \\ 
    MoCov3 \cite{chenEmpiricalStudyTraining2021}
      & \checkmark & \texttimes & \checkmark & \texttimes & \texttimes & \texttimes & \checkmark
      & \texttimes & \texttimes
      & 73.8 & -- & -- \\ 
    OBoW \cite{gidarisOBoWOnlineBagofVisualWords2021}
      & \checkmark & \texttimes & \texttimes & \checkmark & \texttimes & \texttimes & \checkmark
      & \texttimes & \checkmark
      & 73.8 & 61.9 & -- \\ 
    BYOL \cite{grillBootstrapYourOwn2020, caronEmergingPropertiesSelfSupervised2021}
      & \checkmark & \texttimes & \texttimes & \checkmark & \texttimes & \texttimes & \checkmark
      & \texttimes & \checkmark
      & 74.4 & 64.8 & 77.7 \\ 
    DCv2 \cite{caronUnsupervisedLearningVisual2021}
      & \checkmark & \texttimes & \texttimes & \texttimes & \checkmark & \texttimes & \texttimes
      & \texttimes & \checkmark
      & 75.2 & 67.1 & -- \\ 
    SwAV \cite{caronUnsupervisedLearningVisual2021, caronEmergingPropertiesSelfSupervised2021}
      & \checkmark & \texttimes & \texttimes & \texttimes & \checkmark & \texttimes & \texttimes
      & \texttimes & \checkmark
      & \textbf{75.3} & 65.7 & -- \\ 
    DINO \cite{caronEmergingPropertiesSelfSupervised2021}
      & \checkmark & \texttimes & \texttimes & \checkmark & \texttimes & \texttimes & \checkmark
      & \texttimes & \checkmark
      & \textbf{75.3} & \textbf{67.5} & -- \\ 
    \midrule
    \multicolumn{13}{c}{\textbf{ViT-S/16} (21M params, 1007 im/s)} \\
    \midrule
    Supervised \cite{touvronTrainingDataefficientImage2021}
    & -- & -- & -- & -- & -- & -- & --
    & \texttimes & \checkmark
    & 79.8 & 79.8 & 79.8 \\ 
    BYOL \cite{grillBootstrapYourOwn2020, caronEmergingPropertiesSelfSupervised2021}
      & \checkmark & \texttimes & \texttimes & \checkmark & \texttimes & \texttimes & \checkmark
      & \texttimes & \checkmark
      & 71.4 & 66.6 & -- \\ 
    MoCov2 \cite{chenImprovedBaselinesMomentum2020, caronEmergingPropertiesSelfSupervised2021}
      & \checkmark & \texttimes & \checkmark & \texttimes & \texttimes & \texttimes & \checkmark
      & \checkmark & \checkmark
      & 72.7 & 64.4 & -- \\ 
    MoCov3 \cite{chenEmpiricalStudyTraining2021}
      & \checkmark & \texttimes & \checkmark & \texttimes & \texttimes & \texttimes & \checkmark
      & \texttimes & \texttimes
      & 73.4 & -- & 81.4 \\ 
    SwAV \cite{caronUnsupervisedLearningVisual2021, caronEmergingPropertiesSelfSupervised2021}
      & \checkmark & \texttimes & \texttimes & \texttimes & \checkmark & \texttimes & \texttimes
      & \texttimes & \checkmark
      & 73.5 & 66.3 & -- \\ 
    DINO \cite{caronEmergingPropertiesSelfSupervised2021}
      & \checkmark & \texttimes & \texttimes & \checkmark & \texttimes & \texttimes & \checkmark
      & \texttimes & \checkmark
      & \textbf{77.0} & \textbf{74.5} & 82.0 \\ 
    \midrule
    \multicolumn{13}{c}{\textbf{ViT-B/16} (86M params, 312 im/s)} \\
    \midrule
    Supervised \cite{touvronTrainingDataefficientImage2021}
    & -- & -- & -- & -- & -- & -- & --
    & \texttimes & \checkmark
    & 81.8 & 81.8 & 81.8 \\ 
    BEiT \cite{baoBEiTBERTPreTraining2022}
      & \texttimes & \checkmark & \texttimes & \texttimes & \texttimes & \texttimes & \texttimes
      & \texttimes & \checkmark
      & 56.7 & -- & 83.4 \\ 
    MAE \cite{heMaskedAutoencodersAre2021}
      & \texttimes & \checkmark & \texttimes & \texttimes & \texttimes & \texttimes & \texttimes
      & \texttimes & \checkmark
      & 68.0 & -- & \textbf{83.6} \\ 
    MoCov3 \cite{chenEmpiricalStudyTraining2021}
      & \checkmark & \texttimes & \checkmark & \texttimes & \texttimes & \texttimes & \checkmark
      & \texttimes & \texttimes
      & 76.7 & -- & 83.2 \\ 
    DINO \cite{caronEmergingPropertiesSelfSupervised2021}
      & \checkmark & \texttimes & \texttimes & \checkmark & \texttimes & \texttimes & \checkmark
      & \texttimes & \checkmark
      & \textbf{78.2} & \textbf{76.1} & \textbf{83.6} \\ 
    \bottomrule
  \end{tabular}
  \caption{\textbf{Comparison of DINO with previous and concurrent SSL methods on INet-1k}. We report linear probing, kNN, and full fine-tuning performance on the eval split of INet-1k. All networks are trained on INet-1k and inference throughput is measured on a single V100 GPU. Boolean columns indicate for each method whether it is discriminative (Discr), generative (Gen), contrastive (Contr), based on self-distillation (Distil), clustering-based (Clust), based on information maximization (Info), uses a momentum encoder (ME), uses a memory bank (MB), and can tolerate small batch sizes (small BS).}
\end{table*}

\vspace{\subsectionskip}
\subsection{Qualitative Analysis}

\begin{figure}[t]
  \centering
  \includegraphics[width=\linewidth]{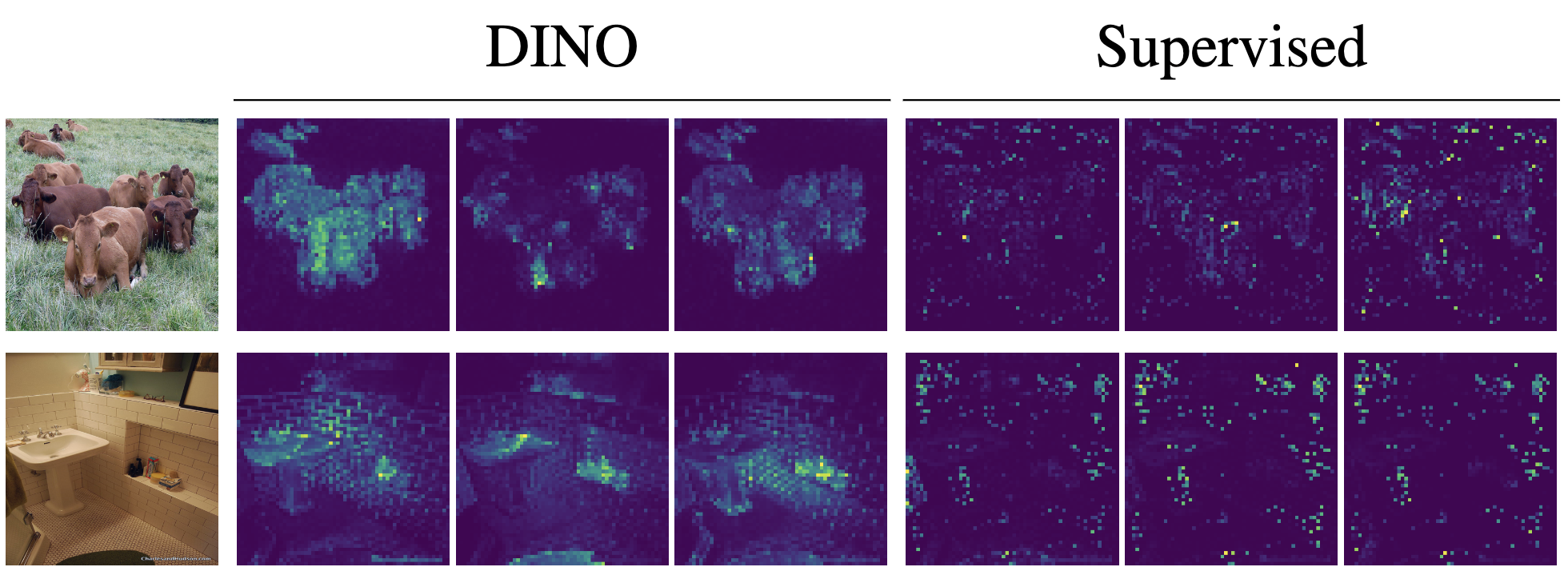}
  \caption{\textbf{Comparison of attention masks between DINO and SL}. The response to the [CLS] token in the last self-attention layer of a ViT is considered. Different columns show different attention heads. Figures from \cite{caronEmergingPropertiesSelfSupervised2021}.}
  \label{fig:dino-attention-comparison-supervised}
\end{figure}

\begin{figure}[h]
  \centering
  \includegraphics[width=\linewidth]{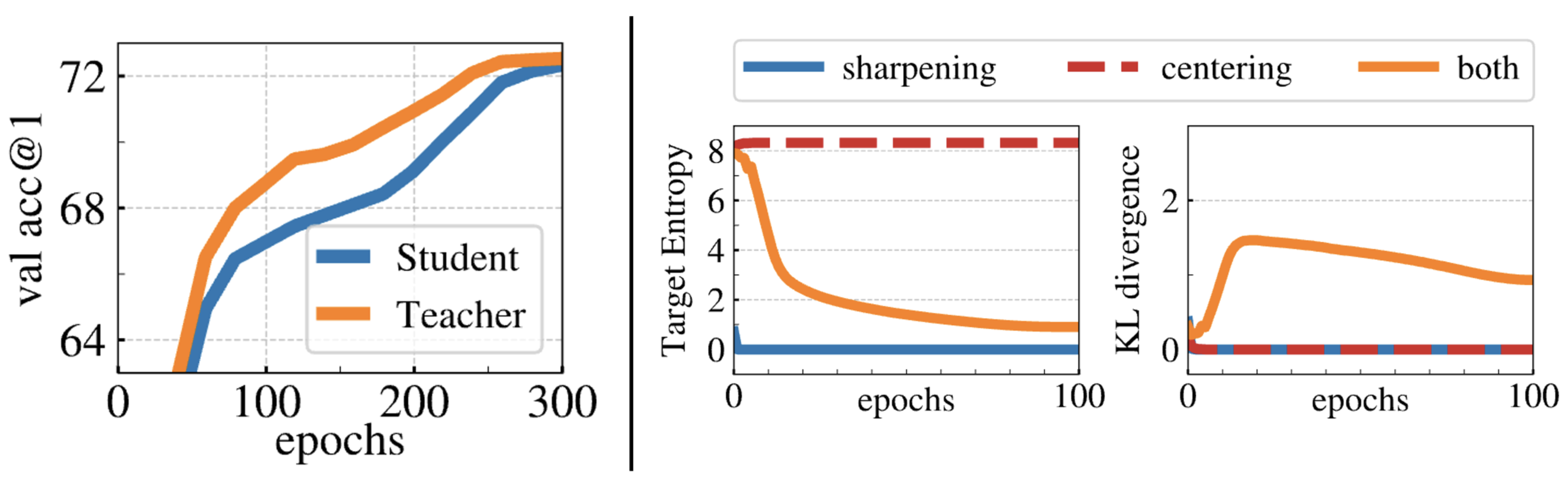}
  \caption{\textbf{Evolution of some metrics during training with DINO}. (Left) Comparison of top-1 accuracy on INet-1k with kNN protocol, using teacher and student frozen embeddings. (Right) Entropy of teacher embeddings and KL divergence between teacher and student embeddings using only centering, only sharpening, or both for teacher targets. Figures from \cite{caronEmergingPropertiesSelfSupervised2021}.}
  \label{fig:dino-diagnostics}
\end{figure}

One important contribution of \cite{caronEmergingPropertiesSelfSupervised2021} is the identification of some interesting properties of the representations learned by Vision Transformers with SSL, which are not found with supervised methods or with convolutional architectures. The remarkably strong off-the-shelf performance in tasks like classification and retrieval, and even in dense tasks like semantic segmentation, has already been discussed in \cref{sec:dino-downstream}. Here, we focus on more qualitative aspects. We also present some findings that shed light on DINO's self-distillation interpretation, and on the mechanism by which it avoids collapse. \\
By analyzing the self-attention patterns of a ViT trained with DINO, the authors of \cite{caronEmergingPropertiesSelfSupervised2021} have shown that the learned features explicitly contain the scene layout and, in particular, object boundaries. These can be recovered by visualizing the response to the query of the [CLS] token in the last self-attention layer, as is depicted in \cref{fig:dino-attention-comparison-supervised}. Interestingly, this is not the case for ViTs trained with supervised learning on the same data. The difference can be quantified by using thresholded self-attention masks to perform a semantic segmentation task, which is shown to work much better with a DINO-trained ViT than with a supervised one (45.9 vs 27.3 Jaccard similarity with ground truth on PASCAL VOC12 \cite{everinghamPascalVisualObject2010} with a ViT-S/16) \cite{caronEmergingPropertiesSelfSupervised2021}. On top of that, with a DINO-trained ViT different heads in the multi-head attention mechanism attend to different semantic regions in the image, even when occluded or small, as is shown in \cref{fig:dino-attention-fine}. Importantly, although this type of information can also be extracted from convolutional networks trained with self-supervision, doing so requires specialized methods \cite{gurVisualizationSupervisedSelfSupervised2020}. \\
To further motivate the interpretation of DINO as self-distillation from a mean teacher, \cref{fig:dino-diagnostics} shows a comparison between the representations of the teacher and student encoders in DINO throughout training, in terms of their downstream performance on INet-1k with a kNN protocol \cite{caronEmergingPropertiesSelfSupervised2021}. The teacher consistently outperforms the student, and its supervisory signal is thus pushing the student to keep developping better representations. Importantly, this is only observed using a mean teacher: if the teacher is built by regularly copying the student weights, the same phenomenon does not happen, and although collapse is avoided the quality of learned representations becomes worse \cite{caronEmergingPropertiesSelfSupervised2021}. \\ 
Finally, to gain insights into the mechanisms by which collapse is avoided in DINO, \cref{fig:dino-diagnostics} shows an ablation in which centering or sharpening are removed, or both. By decomposing the cross-entropy loss between teacher and student embeddings into an entropy term and a KL divergence term, it is shown that centering and sharpening play a complementary role in avoiding collapse (which here means equality between student and teacher embeddings, i.e., KL divergence equal to 0), with the former encouraging uniformity across dimensions, and the latter encouraging a single dimension to dominate \cite{oquabDINOv2LearningRobust2024}.

\vspace{\sectionskip}
\section{Extensions}
\label{sec:extensions}

\begin{table*}
  \centering
  \resizebox{\textwidth}{!}{%
      \begin{tabular}{lllll ccccccccccccc}
          \toprule
          & & & & & \multicolumn{5}{c}{ImageNet [acc.]} & \multicolumn{3}{c}{Classification [acc.]} & \multicolumn{2}{c}{Inst. rec. [mAP]} & \multicolumn{2}{c}{Sem. segm. [mIoU]} & \multicolumn{1}{c}{Depth [RMSE]} \\
          \cmidrule(lr){6-10} \cmidrule(lr){11-13} \cmidrule(lr){14-15} \cmidrule(lr){16-17} \cmidrule{18-18}
          Method & Arch. & \# params & Data & Unsup.
                 & IN-1k & Im-A & Im-R & Im-C [mCE] $\downarrow$ & Sketch
                 & Avg1 & Avg2 & Avg3
                 & Oxford-M & Avg & ADE-20k & Avg & Avg $\downarrow$ \\
          \midrule
          \textbf{SOTA} & - & - & - & \texttimes & 91.0 & - & - & - & - & 1 & 1 & 1 & 90.7 & 1 & 62.9 & 1 & 1 \\
          \midrule
          OpenCLIP \cite{chertiReproducibleScalingLaws2023} & ViT-G/14 & 1843M & LAION-2B & \texttimes & 86.2 & 63.8 & \textbf{87.8} & 45.3 & \textbf{66.4} & 0.85 & \textbf{0.95} & 0.74 & 50.7 & 0.56 & 39.3 & 0.71 & 1.53 \\
          \midrule
          MAE \cite{heMaskedAutoencodersAre2021}     & ViT-H/14 & 632M & INet-1k & \checkmark & 76.6 & 10.2 & 34.4 & 61.4 & 21.9 & 0.47 & 0.81 & 0.56 & 11.7 & 0.10 & 33.3 & 0.65 & 1.53 \\
          DINO \cite{caronEmergingPropertiesSelfSupervised2021}    & ViT-B/8  & 85M & INet-1k & \checkmark & 79.2 & 23.9 & 37.0 & 56.6 & 25.5 & 0.73 & 0.88 & 0.65 & 40.1 & 0.44 & 31.8 & 0.63 & 1.61 \\
          iBOT \cite{zhouIBOTImageBERT2022}    & ViT-L/16 & 307M & INet-22k & \checkmark & 82.3 & 41.5 & 51.0 & 43.9 & 38.5 & 0.79 & 0.90 & 0.72 & 39.0 & 0.48 & 44.6 & 0.79 & 1.29 \\
          \midrule
          \multirow{6}{*}{DINOv2 \cite{oquabDINOv2LearningRobust2024}} 
                   & ViT-S/14 & 21M & LVD-142M & \checkmark & 81.1 & 33.5 & 53.7 & 54.4 & 41.2 & 0.79 & 0.91 & 0.68 & 68.8 & 0.76 & 44.3 & 0.79 & 1.31 \\
                   & ViT-B/14 & 86M & LVD-142M & \checkmark & 84.5 & 55.1 & 63.3 & 42.7 & 50.6 & 0.86 & 0.93 & 0.71 & 72.9 & 0.83 & 47.3 & 0.83 & 1.22 \\
                   & ViT-L/14  & 300M & LVD-142M & \checkmark & 86.3 & 71.3 & 74.4 & 31.5 & 59.3 & 0.89 & 0.93 & 0.73 & \textbf{75.1} & \textbf{0.88} & 47.7 & 0.83 & 1.18 \\
                   & ViT-g/14 & 1100M & LVD-142M & \checkmark & \textbf{86.5} & \textbf{75.9} & 78.8 & \textbf{28.2} & 62.5 & \textbf{0.90} & \textbf{0.95} & \textbf{0.75} & 73.6 & 0.85 & \textbf{49.0} & \textbf{0.84} & \textbf{1.08} \\
                   & ViT-g/14 & 1100M & INet-22k & \checkmark & 85.9 & 73.5 & - & - & - & \textbf{0.90} & - & - & - & - & 46.6 & - & - \\
                   & ViT-g/14 & 1100M & uncurated & \checkmark & 83.3 & 59.4 & - & - & - & 0.81 & - & - & - & - & 48.5 & - & - \\
          \bottomrule
      \end{tabular}
  }
  \caption{\textbf{Comparison of DINOv2 with SSL and WSL alternatives}. We consider a wide range of benchmarks, including both global and dense prediction downstream tasks, with linear probing protocol. Except for INet, we group together similar benchmarks and report a weighted average for each group. Details of the normalization to compute the weighted average, and the exact list of benchmarks considered, can be found in \cref{sec:appendix-benchmarking}. This way, the SOTA is attributed a score of exactly 1, and scores closer to 1 are better. We use an arrow $\downarrow$ to signal that lower is better.}
  \label{tab:vision_models}
\end{table*}


In this section, we discuss two works that built upon DINO's ideas: iBOT \cite{zhouIBOTImageBERT2022}, which introduced a Masked Image Modelling (MIM) objective in the DINO framework, and DINOv2 \cite{oquabDINOv2LearningRobust2024}, which essentially scaled up the approach of iBOT in terms of dataset and model size.

\vspace{\subsectionskip}
\subsection{iBOT}

iBOT \cite{zhouIBOTImageBERT2022} is a method for learning visual representations from images that combines the self-distillation approach from DINO with ideas from the Masked Image Modelling (MIM) literature \cite{baoBEiTBERTPreTraining2022, heMaskedAutoencodersAre2021}. In MIM, part of the input image is masked out, and the model is trained to reconstruct the missing parts. This can be done with a BERT-like architecture \cite{baoBEiTBERTPreTraining2022}, working with a discrete dictionary of patch-tokens obtained through a discrete VAE \cite{rameshZeroShotTexttoImageGeneration2021}, or directly in pixel-space with an autoencoder design \cite{heMaskedAutoencodersAre2021}. \\
The authors of iBOT consider the same joint embedding architecture of DINO, with multi-crop view augmentation, centering and sharpening to avoid collapse, and a mean teacher, and they restrict their attention to transformer backbones (ViT \cite{dosovitskiyIMAGEWORTH16X162021} and SwinT \cite{liuSwinTransformerHierarchical2021}). On top of the self-distillation objective at the image level, they introduce a patch-level MIM objective, using a temperature softmax to normalize patch embeddings and a cross-entropy as reconstruction loss. Given two views \(u\) and \(v\) of an image \(x\), they generate masked views \(\hat{u}\) and \(\hat{v}\) by applying blockwise masking \cite{baoBEiTBERTPreTraining2022}. The student processes masked views, while the teacher processes the original views. The image-level objective matches the [CLS] embeddings across different augmented views; the MIM objective is applied to all pairs of embeddings, within the same augmented view, that correspond to a patch which has been masked out for the student \cite{zhouIBOTImageBERT2022}. \\ 
Compared to DINO, controlling for backbone architecture and pretraining dataset, the addition of the MIM objective in iBOT leads to small but consistent improvements in downstream performance across a variety of tasks, including image classification, object detection, instance segmentation, semantic segmentation, and depth estimation \cite{zhouIBOTImageBERT2022}. Furthermore, it achieves stronger robustness to background change, occlusion, and out-of-distribution examples \cite{zhouIBOTImageBERT2022}. In \cref{tab:vision_models}, we report metrics for iBOT on a variety of benchmarks, comparing with more recent approaches.

\begin{figure}[h]
  \centering
  \includegraphics[width=\linewidth]{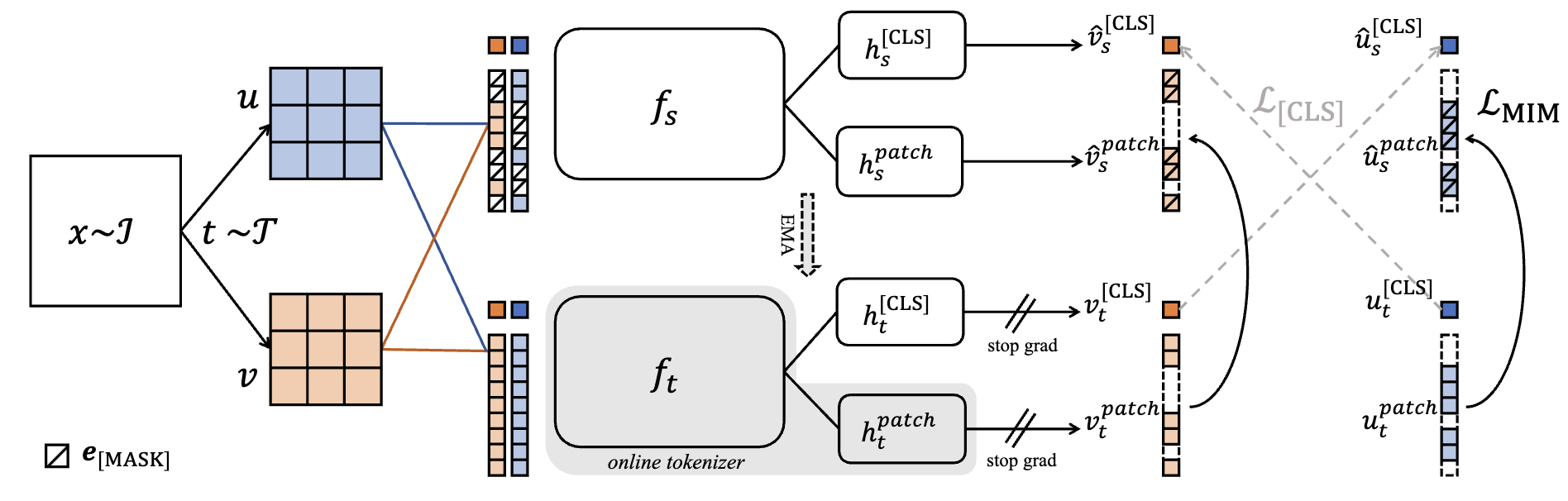}
  \caption{\textbf{Overview of the iBOT framework without multi-crop}. Two views of an image are processed by the teacher, and two masked versions of the same views are processed by the student. The first loss term is self-distillation between cross-view [CLS] tokens, while the second is self-distillation between in-view patch tokens. Figure from \cite{zhouIBOTImageBERT2022}.}
  \label{fig:iBOT}
\end{figure}

\vspace{\subsectionskip}
\subsection{DINOv2}

\begin{figure}[h]
  \centering
  \includegraphics[width=\linewidth]{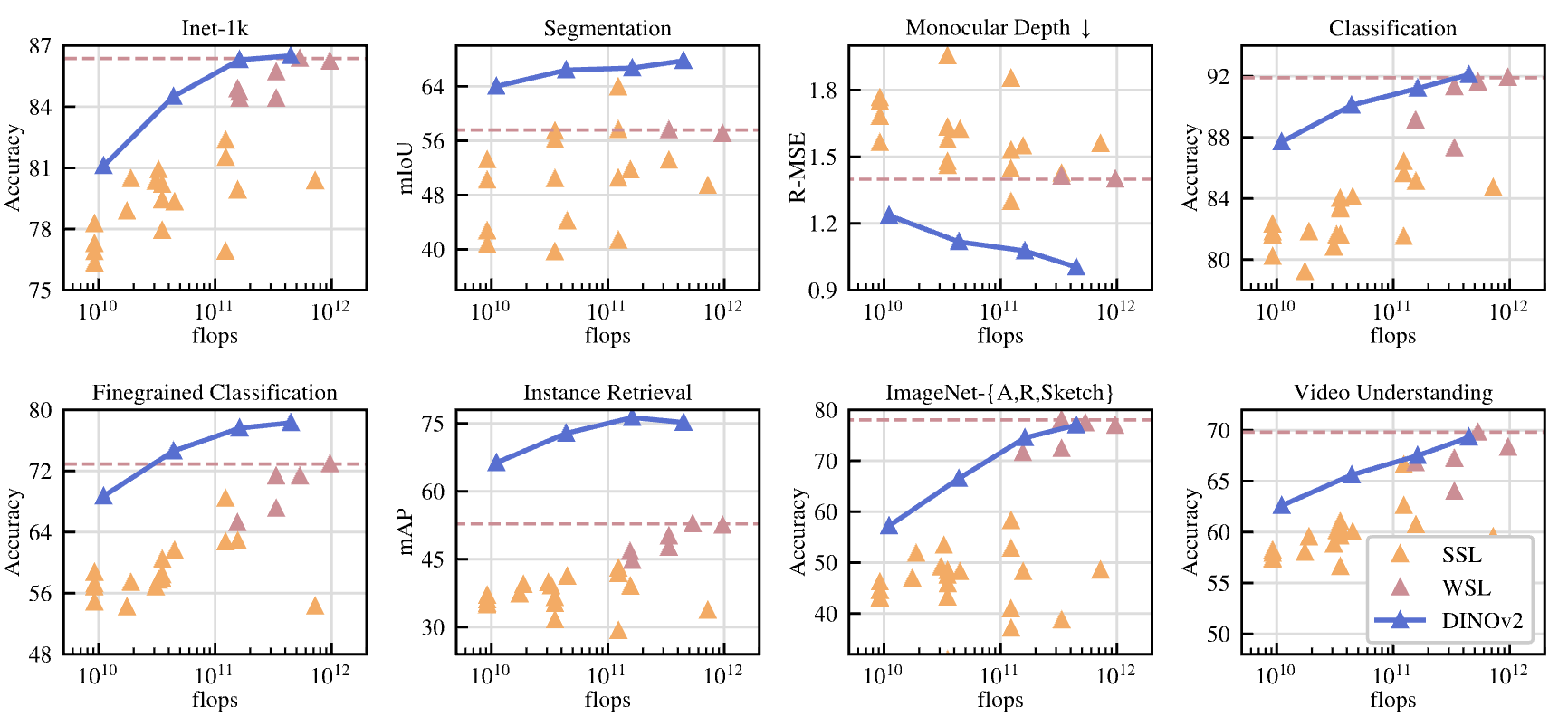}
  \caption{\textbf{Evolution of DINOv2 performance when scaling backbone size}. Performance of four DINOv2 models with increasing backbone size on eight types of vision tasks \cite{oquabDINOv2LearningRobust2024}. The performance of the best self-supervised and weakly supervised methods is shown for comparison. Figure from \cite{oquabDINOv2LearningRobust2024}.}
  \label{fig:dinov2-scaling}
\end{figure}

In DINOv2, the authors propose a series of improvements over iBOT that improve training stability and allow them to effectively scale to larger models and datasets \cite{oquabDINOv2LearningRobust2024}. First, while iBOT uses the same MLP projection head to compute both the image-level and patch-level objective, DINOv2 found that, at scale, these are best learned independently. Second, following \cite{ruanWeightedEnsembleSelfSupervised2023}, DINOv2 substitutes the centering step used in DINO and iBOT before the softmax computation in the teacher with the Sinkhorn-Knopp batch normalization used in SwAV \cite{caronUnsupervisedLearningVisual2021}. Third, they introduced a KoLeo regularizer \cite{sablayrollesSpreadingVectorsSimilarity2019}, applied on l2-normalized features, which encourages a uniform spread of the normalized embeddings of each batch on a hypersphere. Finally, to obtain high downstream performance in pixel-level tasks like segmentation while keeping the pretraining cost under control, they defined a simple curriculum that increases the image resolution during the last phase of training \cite{oquabDINOv2LearningRobust2024}. \\
They also propose an automatic data curation pipeline to retrieve, from a large pool of uncurated data, high quality and deduplicated images that are similar to those in a list of curated datasets \cite{oquabDINOv2LearningRobust2024}. They use it to assemble LVD-142M, a curated dataset of 142M images for the DINOv2 family of models. Their largest model, based on a ViT-g/14 backbone with 1100M parameters, is trained with the DINOv2 algorithm as described above. For smaller models, instead, they use a larger pretrained model from the family as the teacher within the same self-distillation training loop, but without using masking and applying the iBOT loss on the two global crops instead. The final encoder is obtained as an EMA of the student network \cite{oquabDINOv2LearningRobust2024}. \\
In \cref{fig:dinov2-scaling} \cite{oquabDINOv2LearningRobust2024}, we show a comparison between the DINOv2 family of models and the best existing self-supervised and weakly supervised methods on eight types of vision tasks, controlling for FLOPS. DINOv2 matches or surpasses the performance of all other methods, including those using textual supervision, across all tasks and model sizes. In \cref{tab:vision_models}, we collect metrics for the largest versions of DINOv2, OpenCLIP \cite{chertiReproducibleScalingLaws2023}, and some strong SSL methods, including DINO and iBOT, to provide a compact quantitative comparison of their performance. We group similar benchmarks together and report a weighted average of performance on each group, normalizing scores by the inverse of the SOTA performance retrieved from \cite{PapersCodeLatest} or \cite{KaggleYourMachine}. This ensures that all benchmarks have the same importance, despite their scores potentially spanning different ranges. More details on the benchmarks selected and the normalization procedure can be found in \cref{sec:appendix-benchmarking}. From \cref{tab:vision_models}, we can see that the gap with weakly supervised methods is particularly large in dense prediction tasks, like semantic segmentation and depth estimation, which require a fine-grained understanding of the image. On the contrary, in classification benchmarks neither approach demonstrates a consistent advantage. This difference is expected in light of the biases induced by textual supervision. Furthermore, \cref{tab:vision_models} highlights the importance of both size (LVD-142M vs INet-22k) and, even more crucially, quality (LVD-142M vs uncurated) of the pretraining corpus in DINOv2 for dowstream performance.

\vspace{\sectionskip}
\section{Conclusions}
\label{sec:conclusions}

DINOv2 \cite{oquabDINOv2LearningRobust2024} is the first SSL algorithm that was able to produce general-purpose visual features competitive with weakly supervised methods like CLIP \cite{radfordLearningTransferableVisual2021}. This was essentially achieved by scaling up the self-distillation approach of DINO \cite{caronEmergingPropertiesSelfSupervised2021}, with the addition of the Masked Image Modelling objective from iBOT \cite{zhouIBOTImageBERT2022}. The resulting models exhibit strong performance on downstream tasks without any finetuning \cite{oquabDINOv2LearningRobust2024}, including those requiring a fine-graned spatial understanding of images, which enables a wide range of applications with minimal requirements in terms of annotated data and computational resources. Furthermore, their features reveal an emergent understanding of object boundaries and scene layouts without explicit supervision \cite{caronEmergingPropertiesSelfSupervised2021, oquabDINOv2LearningRobust2024}. An interesting direction for future research could be exploring whether scaling model and dataset size even further might lead to more such properties emerging, akin to what has been observed with language models \cite{radfordLanguageModelsAre, brownLanguageModelsAre2020}. Another promising avenue could be integrating some form of textual grounding a posteriori, to enable multimodal applications but leveraging DINO's fine-grained spatial and visual understanding \cite{zhaiLiTZeroShotTransfer2022}. \\
An important limitation of the framework is its reliance on hand-crafted view augmentations. Indeed, this limits the applicability of the method to other modalities beyond images. Works like I-JEPA \cite{assranSelfSupervisedLearningImages2023} explore a similar self-supervised learning approach to DINO, with a joint embedding predictive architecture and a momentum encoder generating targets, but they leverage a simple masking strategy that eliminates the need for hand-crafted augmentations, allowing to easily extend the framework to different modalities \cite{bardesRevisitingFeaturePrediction2024, feiAJEPAJointEmbeddingPredictive2024}. Being able to extract informative and compact representations from any modality is a crucial stepping stone towards the development of general-purpose AI systems, and in particular to learn world models that can be used for learning and planning \cite{lecunPathAutonomousMachinea}. In this direction, DINO's visual understanding has already enabled planning with a learned world model in very simple pixel-based environments \cite{zhouDINOWMWorldModels2024b}, but more work is needed to accommodate complex or multi-modal environments, with applications in sequential decision making, robotics, autonomous driving, and more \cite{yangFoundationModelsDecision2023, guoBYOLExploreExplorationBootstrapped2022, duVideoLanguagePlanning2023, garridoLearningLeveragingWorld2024a, bruceGenieGenerativeInteractive2024, yangLearningInteractiveRealWorld2024a}.

\clearpage
{\small
\bibliographystyle{ieee_fullname}
\bibliography{references}
}

\clearpage
\appendix

\begin{table*}[h]
  \centering
  \resizebox{\textwidth}{!}{%
      \begin{tabular}{lll}
          \toprule
          \textbf{Task} & \textbf{Group Name} & \textbf{Individual Benchmarks} \\
          \midrule
          \multirow{3}{*}{Classification}
          & Avg1 (images)      & iNat18, iNat21, Places205 \\
          \cmidrule(lr){2-3}
          & Avg2 (images)      & Food, Cifar10, Cifar100, SUN, Stanford Cars, Aircr, VOC, DTD, Pets, Cal101, Flowers, CUB \\
          \cmidrule(lr){2-3}
          & Avg3 (videos)      & K400, UCF-101, SSv2 \\
          \midrule
          Instance Recognition & Avg & Oxford M, Oxford H, Paris M, Paris H, AmsterTime \\
          \midrule
          Semantic Segmentation & Avg & ADE20k, CityScapes, Pascal VOC \\
          \midrule
          Depth Estimation & Avg & NYUd, KITTI, NYUd $\rightarrow$ SUN RGB-D \\
          \bottomrule
      \end{tabular}
  }
  \caption{\textbf{Composition of benchmark groups considered in \cref{tab:vision_models}.}}
  \label{tab:appendix-benchmarking}
\end{table*}

\section{Benchmarking Details}
\label{sec:appendix-benchmarking}

Here, we provide more details about the exact benchmarks that make up each group in \cref{tab:vision_models}, and the normalization procedure used to compute the weighted average scores. \\
The weighted average is designed to give equal importance to each benchmark, regardless of its typical score range. Indeed, with naive averaging, benchmarks with a larger variance in their typical scores would have a larger impact on the result. To avoid this, for each benchmark considered, we retrieve the SOTA performance from \cite{PapersCodeLatest} or \cite{KaggleYourMachine}, and we normalize the scores obtained by each method by the reciprocal of the SOTA score. This way, the SOTA is attributed a score of exactly 1, and scores closer to 1 are better. We then take a simple arithmetic mean of the normalized scores to obtain the weighted average for a group of benchmarks. \\
\cref{tab:appendix-benchmarking} describes the exact composition of each benchmark group in \cref{tab:vision_models}.

\end{document}